\documentclass[a4paper]{article}

\usepackage{INTERSPEECH2020}

\usepackage{xcolor}
\usepackage{makecell}

\usepackage{hyperref}
\usepackage{xstring}
\usepackage{soul}

\newcommand{\et}[1]{{\bf [\textcolor{blue}{EDMONDO #1}{\bf ]}}}

\newcommand{\sap}[1]{{\bf [\textcolor{violet}{SARA #1}{\bf ]}}}

\newcommand{\fbkb}{FbkB}

\newcommand{\COMMENT}[1]{}
\usepackage{acronym}

\newacro{HMMs}[HMMs]{hidden Markov models}
\newacro{CALL}[CALL]{computer assisted language learning}
\newacro{MLLR}[MLLR]{maximum likelihood linear regression}
\newacro{GMM}[GMM]{gaussian mixture model}
\newacro{WER}[WER]{word error rate}
\newacro{ASR}[ASR]{automatic speech recognition}
\newacro{DAE}[DAE]{denoising autoencoder}
\newacro{DNN}[DNN]{deep neural network}
\newacro{BN}[BN]{batch normalization}
\newacro{RNN}[RNN]{recurrent neural network}
\newacro{CNN}[CNN]{convolutional neural network}
\newacro{RBM}[RBM]{restricted boltzmann machine}
\newacro{L2}[L2]{second language}
\newacro{MSE}[MSE]{mean square error}
\newacro{LM}[LM]{language model}
\newacro{B-LSTM}[B-LSTM]{bidirectional long-short term memory}
\newacro{SAT}[SAT]{speaker adaptive training}

\title{Mixtures of Deep Neural Experts for Automated Speech Scoring}
\name{Sara Papi$^1$, Edmondo Trentin$^1$, Roberto Gretter$^2$, Marco Matassoni$^2$, Daniele Falavigna$^2$ }
\address{
  $^1$DIISM - Universit\`a di Siena, Italy\\
  $^2$Fondazione Bruno Kessler (FBK), Italy}
\email{sara.papi@student.unisi.it, trentin@dii.unisi.it, (gretter,matasso,falavi)@fbk.eu}

\begin{document}

\maketitle
\begin{abstract}
The paper copes with the task of automatic assessment of second language proficiency from the language learners' spoken responses to test prompts. The task has significant relevance to the field of computer assisted language learning. The approach presented in the paper relies on two separate modules: (1) an automatic speech recognition system that yields text transcripts of the spoken interactions involved, and (2) a multiple classifier system based on deep learners that ranks the transcripts into proficiency classes. Different deep neural network architectures (both feed-forward and recurrent) are specialized over diverse representations of the texts in terms of: a reference grammar, the outcome of probabilistic language models, several word embeddings, and two bag-of-word models. Combination of the individual classifiers is realized either via a  probabilistic pseudo-joint model, or via a neural mixture of experts. Using the data of the third Spoken CALL Shared Task challenge, the highest values to date were obtained in terms of three popular evaluation metrics.
\end{abstract}

\section{Introduction}
\label{itroduction}

The problem of automatic proficiency scoring in second language learning (L2) \cite{zechner2009} has been largely investigated in the past in the framework of computer assisted language learning (CALL) \cite{garrett2009}. Approaches have been proposed for two input modalities: written and spoken. In both cases, specific competencies of the human learners are processed by suitable proficiency classifiers. The goal is to 
measure L2 proficiency relying on some ground truth provided by human experts. To this aim, the paper proposes and investigates the use of models for proficiency classification on a public data set released for the
third Spoken CALL Shared Task \cite{Baur2019} challenge.  This  took place in 2019 (hereafter ``2019 challenge''), and addressed the automatic scoring of sentences uttered by Swiss German teenagers learning English in their second and third year. 
\COMMENT{
\sap{questo viene spiegato anche sotto, lo toglierei da qui}Sentences are responses in English to written German prompts. The task consists  in  assigning to each spoken response a binary label, ``accept" or ``reject", trying to match, as closely as possible, an unknown gold standard. The utterance is accepted if it is correct in both grammar and meaning, otherwise it is rejected. \sap{fino a qui, in generale nell'introduzione accennerei solo che il dataset è stato usato per il proficiency scoring e citerei i vari tentativi che sono già stati fatti, come avete già fatto qui sotto}
}

Most of the approaches used by participants in the 2017 and 2018 editions of the challenge \cite{Baur2017,Baur2018} rely on hand-crafted features, extracted from both audio signals and automatic transcriptions of utterances, fed to a traditional classifier (e.g., based on logistic regression). These approaches, used also in most commercial systems (see \cite{zechner2019} for a review), exhibited good performance on the task of the challenge.  In both 2018 and 2019 challenges \cite{Baur2018, Baur2019} some approaches based on word embeddings \cite{mikolov2013} were investigated, as well.

In the work reported in \cite{russell2019}, sentence similarities among ASR transcripts and the corresponding responses contained in a reference  ``grammar'' (i.e.\ a non-exhaustive, prompt-specific list of appropriate responses, provided by the organizers of the challenge) are used as features for a scoring system based on a neural network (NN). The performances obtained using several in-domain and out-of-domain word embeddings (namely, Word2Vec and doc2vec \cite{mikolov2014}) are compared in \cite{russell2019}, as well. An alternate approach based on word embeddings was presented in \cite{sokhatskyi2019} for the 2019 challenge. The scoring system proposed in \cite{sokhatskyi2019} relies on a NN fed with 918-dimensional word vectors. Each vector is formed by concatenating the outputs of the Bidirectional Encoder Representations from Transformers (BERT) \cite{Devlin2019} and a NN-based language model (LM) \cite{bengio2003} trained on the data sets delivered for the challenge. Also, alternative word embeddings (i.e. word2vec, doc2vec, and ELMO \cite{peters2018}) were evaluated experimentally in \cite{sokhatskyi2019}. 

In the present paper, building on the aforementioned experiences, we propose an approach that properly combines the outputs of several scoring systems, including the system winner of the 2019 challenge \cite{gretter2019}. A speech recognizer is applied first, so as to obtain transcripts of the  noisy responses uttered by the students. Feedforward and recurrent deep neural networks (DNN) are then used to model different representations of the automatic transcriptions. For each of these representations, a DNN is trained (hence, specialized) over a corresponding set of features, namely: (i) the scores yielded by a reference grammar, (ii) the likelihoods estimated by a number of probabilistic LMs, (iii) sequences of word embeddings of different type, and (iv) two variants of the bag-of-words representation. Besides applying individually each of these DNN-based ``experts'', multiple classifier systems are presented that combine all of them into a higher-level, more robust classifier capable of exploiting the specific capabilities of the individual experts. The combination is accomplished by either applying a pseudo-joint probability criterion over the individual DNN estimates \cite{TrentinLC11}, or by means of a mixture of DNNs~\cite{surveyME2014}. In so doing the highest values to date are obtained in terms of the evaluation metric used officially for the challenge, as well as of other popular metrics. 

\COMMENT{
Scientific literature contains several  papers describing the usage of word embeddings in the field of CALL application. In the following we will summarize those that we believe are more related to the approaches proposed in this paper. 

In the work reported in \cite{QianSLT18} an attention based long short-term memory (LSMT) recurrent NN (RNN) was proposed for scoring  spoken answers by non native speakers. Word embeddings of input utterances are first computed from their related automatic transcriptions,  then they are concatenated with an encoded representation (also derived from a bidirectional LSTM network) of the corresponding prompts and processed by the attention based LSTM.  This  {\em prompt aware} conditioning  mechanism allows to achieve performance similar to that of a traditional content based scoring system  which, however, makes use use of both hand-crafted features and prompted dependent models.

%\df{citare papers di bernstein and pearson}

There are several papers comparing NN-based text scoring  with features based text scoring \cite{riordan2017,alikaniotis2016,dong2016}. In principle the proposed approaches  could be also applied for scoring spoken utterances after having automatically transcribed them. The work in \cite{alikaniotis2016} uses and end-to-end model for essay scoring based on word embeddings trained on essay scores. This model achieves on a well known data set better performance than a standard model based on hand-crafted features.
}

%{\bf Related works.}

The combination of multiple DNNs for speech scoring was proposed in \cite{QianICASSP18}, where two DNNs are employed to encode the lexical and the acoustic information contained in a spoken utterance, respectively. The lexical DNN encodes an automatically recognized input sentence relying on a pre-trained model (namely ``GloVe", described in \cite{glove2014}), while the acoustic DNN encodes a corresponding sequence of word-level acoustic features. A linear regression model is used to combine the scores provided by the two DNNs.  Different DNN architectures were evaluated empirically in \cite{QianICASSP18}. The best results were obtained via a bidirectional LSTM (long-short term memory) \cite{HochSchm97} NN, together with an attention mechanism. 

Mixtures of experts (in the form of shallow NNs) have long been investigated and successfully applied to a number of tasks in the machine learning community \cite{surveyME2014}. Applications of mixtures of experts include acoustic modeling \cite{AcousticModelsME}, language modeling and machine translation \cite{ShazeerMMDLHD17}, and other natural language processing tasks \cite{LeDR16}. Since the outbreak of deep learning \cite{ShazeerMMDLHD17}, also mixtures of DNNs have been receiving increasing attention. We follow in the footsteps of \cite{HardMixture}, insofar that a hard mixture of DNNs is built from independent experts that are individually specialized on expert-specific features, and efficiently trained in parallel. A probabilistic gating function can be applied that realizes a pseudo-joint likelihood of the independent DNNs (provided that a proper probabilistic interpretation of the DNNs outputs is given). More generally, a higher-level gating network can be trained a posteriori to assign individual credit to the pre-trained experts.

%\section{The third spoken CALL shared task}
%\section{Task and baseline system}
\section{Task and systems description}
\label{sec:sharedtask}

%\mm{Tagliare e fondere con section ASR system}

The third Spoken CALL Shared Task \cite{Baur2019} is composed by {\em(prompt, response)} pairs where prompts are written questions in German, while responses are speech recordings of spoken utterances given in English by native German-speaking Swiss teenagers.
Each response was tagged by human raters with two Boolean labels denoting the correctness of the responses in terms of {\em language} and {\em meaning}, respectively. The task consists in classifying utterances as {\em accepted} or {\em rejected}: a response shall be accepted if both its {\em language} and its {\em meaning} are labeled as ``correct''. A reference grammar  is made available by the organizers of the challenge, in the form of a list of correct written responses to each given prompt.  % In the past editions of the challenge, the organizers produced two training sets and two evaluation sets.
 %Each pair normally comes with additional information, namely: an {\em Id}, an orthographic transcription, the output of an ASR baseline, 
  %Part of the training data, that have been semi-automatically labelled, contains also  information that summarizes the labelling process of the corresponding utterances (field {\it Trace}), not used in this work.  
For the 2019 challenge, the two training sets of the past editions~\cite{Baur2017,Baur2018} are merged to form the training set (hereinafter called {\em TrainingSet}, 11919 utterances), the test set of the first challenge plays the role of development set ({\em DevSet}, 995 utterances), and the test set of the second edition is adopted as evaluation set ({\em EvalSet}, 1000 utterances). {\em TrainingSet} was used to train the acoustic models and the LMs. {\em DevSet} and {\em EvalSet} were used to tune hyper-parameters of the whole system and for applying model selection,
%to measure the corresponding performance
respectively. Finally, the test data of the 2019 challenge ({\em TestSet}, 1000 utterances) were used for testing the performance of the resulting system.

%   TT3ScAsrM1/scst3_dev_IPWTRLM.csv            995
%   TT3ScAsrM1/scst3_eval_IPWTRLM.csv          1000
%   TT3ScAsrM1/scst3_train_IPWTRLMT.csv       11919
%   TT3ScAsrM1/scst3_test_IPWTRLM.csv          1000

\COMMENT{
\begin{table}[h]  \centering
\begin{tabular}{ c c c }  \hline
 Id         & \# of Utterances & Source   \\ \hline 
DevSet      &   995      & scst1 Test \\
EvalSet     &  1000      & scst2 Test \\
TrainingSet & 11919      & scst1 Train + scst2 Train \\ \hline
TestSet     &  1000      & scst3 Test \\ \hline
\end{tabular} \caption{Data sets of the challenge used in this paper.}
\label{tab:sets} \end{table}
}

%\section{ASR system}
%\label{sec:ASR}
%\mm{tagliare e fondere con section 3}

%\subsection{Acoustic model}

The acoustic model, improved over \cite{matassoni2018}, was trained using a popular Kaldi recipe \cite{povey2016} that relies on a time-delay NN optimized using the lattice-free maximum mutual information approach, i.e.\ with a sequence-level objective function. The acoustic model was trained on an extended dataset that, in addition to {\em TrainingSet}, embraced (i) the subset of {PF-STAR} \cite{batliner2005} comprising the recordings of read English speech from German children (about 3.5h), and (ii) the {ISLE} corpus \cite{menzel2000isle}, consisting of 11484 utterances recorded by intermediate-level German learners of English (about 18h). As for the LM, the 3-gram stochastic LM provided by the organizers was adopted (details in~\cite{Baur2019}), resulting in a 7.5\% word error rate on {\em EvalSet}.

%We have also trained a new language model using the manual transcriptions of the full TrainingSet and obtained a second ASR system that performs slightly better in terms of WER.

%\subsection{ASR performance}
%\label{subsec:asrperformance}

%Table~\ref{table:rec_results} reports the main features of the adopted ASR system and the resulting WERs on the EvalSet.

%\begin{table}[h]
%\begin{tabular}{ l c c c }
% & AM & LM & WER  \\ \hline
%AsrV1 & \makecell{TDNN, LF-MMI, train on %\\TrainingSet + PF-STAR + ISLE} & %\makecell{3-grams \\ scst1}  & 7.6  \\
%AsrV2 & \makecell{TDNN, LF-MMI, train on %\\TrainingSet + PF-STAR + ISLE} & %\makecell{3-grams \\ scst1+2}  & 7.5  \\
%\hline
%\end{tabular}
%\caption{WER results on the EvalSet, provided by our ASR systems.}
%\label{table:rec_results}
%\end{table}
%\vspace{-0.8cm}

\COMMENT{
\section{Proposed approach}
\label{sec:proposapproach}
\et{}
Given the textual transcriptions obtained from the speech signals, the task was formalized in terms of a 4-class classification problem, depending on the grammatical and semantic correctness/incorrectness of each assessee's answer provided during the test. Specific classifiers (i.e., ``experts'') were specialized over different sets of features extracted form the texts, as follows.
}

\subsection{FBK baseline system}
\label{subsec:FBKsys}

The winner of the 2019 challenge
 used the following sets of features:
 {\it standard}, 4 features counting the number of words, of content words, the number and percentage of out-of-vocabulary (OOV) words;
{\it reference}, 5 features computed using the reference grammar and the edit error (see ~\cite{gretter2019});
{\it LMs}, features computed using some LMs. For each LM, 5 features related to log-probability, OOVs and number of back-offs were computed. A maximum of 12 LMs were defined, 1-grams to 4-grams, computed on 3 data sets: {\it Generic}, around 3 million words from English TED talks; {\it TrainRejRec}, ASR outputs bounded by labels {\it \_start\_} and {\it \_end\_}, corresponding to the incorrect utterances of {\em TrainingSet}; {\it TrainAccRec}, the same but corresponding to the correct utterances of {\em TrainingSet}.

Several feed-forward NNs (FFNNs) were used to perform classification; then, majority voting was applied to the most promising (on {\em DevSet}) classification outputs to contrast the high variability of the results observed on {\em DevSet}.

\COMMENT{
The winner of the 2019 Spoken CALL Shared Task
 used the following set of features:
\begin{itemize}
\item {\bf standard}, 4 features, namely: number of automatically recognized words, number of content words, number of out-of-vocabulary (OOV) words, percentage of OOV;
\item {\bf reference}, 5 features computed using the reference grammar and the edit error (see [slate2019] for details);
\item {\bf LMs}, features computed by means of some LMs. For each LM, 5 features related to log-probability, OOVs and number of back-offs were computed. A maximum of 12 LMs were defined: 1-grams to 4-grams computed on 3 data sets:
$\bullet$ {\bf Generic}: around 3 millions of words belonging to transcriptions of English TED talks;
$\bullet$ {\bf TrainRejRec}: ASR outputs, bounded by labels {\it \_start\_} and {\it \_end\_}, corresponding to the incorrect utterances of TrainingSet;
$\bullet$ {\bf TrainAccRec}: the same, but corresponding to the correct utterances of TrainingSet.
\end{itemize}

Simple feed-forward \hl{NNs} were used to perform classification, varying some meta-parameters. Then, a majority voting was applied to the most promising (on the DevSet) classification outputs to contrast the high variability of the results, observed on the DevSet.
}

\subsection{Stand-alone DNNs and textual features used}
\label{sec:DNNs}

Sections \ref{subsec:LSTM_W2V} and \ref{subsec:MLP_BOW} present the stand-alone DNNs used in the present paper, along with the corresponding features extracted from the prompts and the transcripts of the responses.  Unless otherwise stated, henceforth a categorical cross-entropy loss function is used for training the networks.

\subsubsection{LSTM on word embeddings (Word2Vec, BERT)}
\label{subsec:LSTM_W2V}

The LSTM model is considered first, and trained over sequences of 300-dimensional real-valued Word2Vec \cite{mikolov2013} word embeddings (hereafter we write LSTM-W2V to refer to this approach). The Word2Vec embeddings for any given {\em (prompt, response)} pair are then concatenated in order to form a single sequence of vectors, corresponding to the words in the prompt followed by the words in the response.

An ad hoc version of the loss function to train the LSTM-W2V is devised, as well, so as to account for the mismatch between the usual training criterion and the evaluation metric used in the 2019 challenge, that is the $D_{full}$ \cite{Baur2018}. The following criterion is proposed:

\begin{equation*}
        L(y,\hat y)= \begin{cases} \lambda MSE(y,\hat y), & \mbox{if } y\mbox{ is correct in language and }\\ & \hat y\mbox{ is incorrect in meaning} \\
        MSE(y,\hat y), & \mbox{otherwise} \end{cases}
\end{equation*}
    
\noindent where MSE denotes the mean squared error loss, whose value is multiplied by $\lambda$ (where $\lambda > 1$) whenever the meaning of the current response is incorrect while the system accepts it (thus, penalizing the {\em gross false accept} errors). We write LSTM-W2V-L to refer to the LSTM trained over the modified loss. Within the experimental framework reported in Section \ref{sec:exp}, a grid-search model selection procedure singled out an empirically suitable value of $\lambda = 3$. 
    
Another variant of LSTM-W2V is achieved by interposing an end-of-sentence marker amidst the sequence of embeddings representing the prompt and the sequence representing the corresponding response. Roughly speaking, the marker is expected to provide the network with explicit information on when exactly to switch its internal state from prompt-processing to response-processing, possibly easing the LSTM learning and classification tasks. Henceforth, the approach is referred to as LSTM-W2V-M. In practice, the present variant turns out to yield the best performance when realized by means of an additional component to the embedding vectors (which become 301-dimensional), where the additional component is permanently set to zero except for the aforementioned marker, where it is set to one (other components in the marker are set to zero).

Finally, we replaced Word2Vec with BERT embeddings \cite{Devlin2019} (768-dimensional real-valued vectors)  applying the same concatenation procedure of the word encodings in the prompt and in the response, respectively, separated by a marker (to this end, an additional binary component was added to the BERT vectors, as well, resulting in a 769-dimensional feature space).  The present variant is abbreviated as LSTM-BERT.

Note that the use we propose of LSTMs with sequences of Word2Vec or BERT vectors is different from the use made in \cite{russell2019}. The latter 
relies on Word2Vec and doc2vec embeddings in order to compute individual sentence similarities between the responses and the reference grammar for the challenge. Such similarities are then used to train FFNN-based classifiers. In like manner, our approach differs entirely from that proposed in \cite{sokhatskyi2019}. In fact, the latter averages over the sequence of  embeddings in the prompt and in the corresponding response, obtaining a single static vector, which is concatenated with another vector drawn from a probabilistic LM. This flat, fixed-dimensional representation is eventually fed into a shallow FFNN-based classifier \cite{sokhatskyi2019}. %In Section \ref{sec:exp} we also compare empirically our approaches with the best-performing solutions presented by \cite{russell2019} and \cite{sokhatskyi2019}.

\subsubsection{Deep FFNN on Bag-Of-Words and TF-IDF}
\label{subsec:MLP_BOW}

A deep FFNN was applied to two completely different encodings of the transcripts, obtained by extracting either bag-of-words (BOW) or term frequency-inverse document frequency (TF-IDF) \cite{2017Goldberg} vectors from the prompt and the response, respectively. The BOW model considered herein is a plain counter of the occurrences of individual words in the text. Both BOW and TF-IDF representations rely on a 1020 word vocabulary (embracing both German prompts and English responses in the dataset).  The 2040-dimensional input vector fed to the FFNN was formed by concatenation of the encodings corresponding to the current prompt and response. %Model selection ended up prescribing a 5-layer FFNN (150, 150, 150, 170, and 15 neurons per layer) for the BOW-based encoding, as well as a 5-layer FFNN (210, 130, 150, 170, and 15 neurons per layer) for the TF-IDF.

\subsection{Combining multiple DNNs}
\label{subsec:MoE}

A first probabilistic strategy for combining the different DNNs is readily achieved in terms of a pseudo-joint probability model as follows \cite{TrentinLC11}. Henceforth, we write $\omega_1, \ldots, \omega_c$ to represent the $c$ different classes involved in the classification task. Let $\psi_1$ and $\psi_2$ be the continuous-valued random quantities yielded by two distinct functions (or, regression models) of the outcome ${\bf x}$ of a given random phenomenon (e.g., the transcript of the response to a prompt). We refer to $\psi_1$ and $\psi_2$ as the ``models'' (it is straightforward to extend the following discussion to an arbitrary number of models). Different representations of ${\bf x}$ are given in terms of model-specific random vectors of features (or, sequences of feature vectors), say ${\bf x}^{(1)}$ for $\psi_1$ and ${\bf x}^{(2)}$ for $\psi_2$, respectively. Assuming that the models are independent of each other, for any state of nature $\omega_i$ ($i=1, \dots, c$) we can write:

\begin{eqnarray}
\label{eq:models}
P(\omega_i \mid  \psi_1, \psi_2) & = & \frac{p(\psi_1, \psi_2 \mid \omega_i)P(\omega_i)}{p(\psi_1, \psi_2)}\\ \nonumber
 & = & \frac{p(\psi_1 \mid \omega_i)p(\psi_2 \mid \omega_i)P(\omega_i)}{p(\psi_1) p(\psi_2)}\\ \nonumber
% & = & \frac{P(\omega_i \mid \psi_1)}{P(\omega_i)} \frac{P(\omega_i \mid \psi_2)}{P(\omega_i)} P(\omega_i)\\ \nonumber
 & = & \frac{P(\omega_i \mid \psi_1)P(\omega_i \mid \psi_2)}{P(\omega_i)}
\end{eqnarray}

\noindent where Bayes theorem was used in the third step of the calculations. Equation (\ref{eq:models}) allows the computation of $P(\omega_i \mid  \psi_1, \psi_2)$ is terms of a pseudo-joint probability (the product of quantities at the numerator) normalized by the class-prior. The rationale behind the use of the expression ``pseudo'' lies in the fact that, in general, the models are actually not independent of each other under real-world circumstances. Nonetheless, equation (\ref{eq:models}) can still be applied, for practical intents and purposes, in a naive-Bayes fashion. A discriminant function $g_i(.)$ can thus be defined for each class $\omega_i$ by revolving around the usual maximum-a-posteriori probability given the models, i.e.\ $\max_i P(\omega_i \mid  \psi_1, \psi_2)$. In the light of equation (\ref{eq:models}) such a discriminant function turns out to be defined as a normalized pseudo-joint probability in the form $g_i({\bf x}) =  P(\omega_i \mid \psi_1({\bf x}^{(1)}))P(\omega_i \mid \psi_2({\bf x}^{(2)}))/P(\omega_i)$, and the resulting decision rule assigns ${\bf x}$ to class $\omega_i$ if  $g_i({\bf x}) \geq g_j({\bf x})$ for each $j \neq i$, as usual. Hereafter we assume that $\psi_1(.)$ and $\psi_2(.)$ are the functions computed by two distinct stand-alone, feature-specific DNNs to be combined, and we let $P(\omega_i \mid\psi_j({\bf x}^{(j)})) \approx  \psi_j({\bf x}^{(j)})$ in compliance with the formal probabilistic interpretation of the DNN outputs as estimates of the class-posterior probabilities \cite{TrentinF09}. It is seen that equation (\ref{eq:models}) can be readily extended to any number $k$ of models $\psi_1, \ldots, \psi_k$ with model-specific representations ${\bf x}^{(1)}, \ldots, {\bf x}^{(k)}$ of ${\bf x}$, allowing for a pseudo-joint combination of an arbitrary number of feature-specific DNNs.
 
The second technique for combining the stand-alone DNNs described in the previous sections relies on a hard mixture of experts \cite{HardMixture}. While in \cite{HardMixture} the mixture is hard insofar that the individual experts are trained independently over a crisp partitioning into expert-specific clusters of the (shared) feature space, herein the overall set of available features is partitioned into homogeneous, non-overlapping subsets of specialized features, and each DNN expert takes (independently) responsibility for the corresponding feature-specific representation of the {\em (prompt, response)} pairs of the whole dataset. In so doing, the stand-alone DNNs presented in the previous sections can be used as the (pre-trained) experts. The different DNNs in the mixture are then combined using a gating network, as follows. Assuming that $k$ experts $E_1, E_2, \ldots, E_k$ are involved in the mixture, let ${\bf x}^{(j)}$ represent the specific feature vector (or, sequence of feature vectors) for the generic $j$-th neural expert $E_j$ (i.e.\ ${\bf x}^{(j)}$ may represent the BERT-based embeddings of current {\em (prompt, response)} pair, or the corresponding BOW-based representation, etc.). Let ${\bf y}_j = \varphi_j({\bf x}^{(j)})$ be the function computed by $E_j$ over ${\bf x}^{(j)}$, and let ${\bf y} = ({\bf y}_1, \dots, {\bf y}_k)$ be the vector embracing all the experts outputs. The gating network is trained to compute a mapping between its input vector ${\bf y}$ and a $k$-dimensional credit vector $(\alpha_1({\bf y}), \ldots,  \alpha_k({\bf y}))$ such that the overall output ${\bf z}$ of the mixture is defined as ${\bf z}=\sum_{j=1}^{k}\alpha_j({\bf y}) \varphi_j ({\bf x}^{(j)})$, where $\alpha_j({\bf y}) \in (0,1)$ is the credit assigned by the gating network to the $j$-th expert, for $j=1, \ldots, k$. In so doing, no arbitrary prior choices are imposed on the overall multiple-classifier combination criterion: in fact, the  machine learns from the available examples how to assign credit to the individual experts of the mixture. The gating network is trained over the modified criterion function $L(\cdot,\cdot)$ presented in Section \ref{subsec:LSTM_W2V}. It is seen that defining the mixture this way allows for combining both FFNN experts and recurrent LSTM experts within the same framework. The proposed mixture generalizes the notion of linear regression model over the regressors ${\bf y}_1, \ldots, {\bf y}_k$, insofar that the regression parameters $\alpha_1({\bf y}), \ldots,  \alpha_k({\bf y})$ are themselves parametric nonlinear functions of the regressors themselves.

%Note that each $\alpha_j$ may or may not be an explicit function of ${\bf x}$, depending on whether the latter is include din the input vector fed into the gating network (both variants are investigated empirically in the next Section).

\section{Experiments and results}
\label{sec:exp}

The \emph{TrainingSet}, \emph{DevSet} and \emph{EvalSet} were first merged and then, applying a uniform random sampling, partitioned into two subsets: the training set (80\% of the data) and the validation set (20\% of the data). These subsets were used, respectively, to train the DNNs and to accomplish model selection \cite{ANDERS1999309}, respectively. The selected models were eventually evaluated in terms of different metrics on {\em TestSet}. Both the popular {\em Adam} and {\em RMSprop} optimizers were applied for training the DNNs. A grid-search model selection procedure was used \cite{BergstraB12}, which ended up prescribing a learning rate equal to 0.001 for all models except for the MIX2-3\fbkb-F1 variant (see below), where a learning rate of 0.0001 was selected. As for the number of training epochs, we relied on the early stopping strategy based on the validation loss. Early stopping resulted in an overall number of epochs in the range $[30\div 50]$, depending on the DNN and on the features used. \autoref{tab:architectures} reports the architectures (number of layers and number of neurons per layer, in the order) of the stand-alone DNNs as determined via model selection.

\begin{table} [h]
\begin{center}
\begin{tabular}{c c c}
\hline
{Neural Network} & Layers & Neurons per layer\\
\hline
LSTM-W2V & 3  & 300-300-4\\
LSTM-W2V-L & 3  & 300-175-4\\
LSTM-W2V-M & 6 & 301-75-25-25-20-4\\
FFN-BOW-WC & 7  & 2040-150-150-150-170-15-4 \\
FFN-BOW-TFIDF & 7 & 2040-210-130-150-170-15-4 \\
LSTM-BERT & 6 & 769-100-100-20-20-4  \\
\hline
\end{tabular}
\caption {DNNs: no. of layers and no. of neurons per layer.}
\label{tab:architectures}
\end{center}
\vspace{-1 cm}
\end{table}

The metrics used are the $D_{full}$ (the official metric of the 2019 challenge \cite{Baur2018}), the accuracy (percentage of correct classifications), and the F-measure ($F1$). The test results in terms of $D_{full}$ obtained from the selected stand-alone DNNs are shown in the second column of  \autoref{tab:proposed_approaches_results}, where they are compared with the baseline performance yielded by the 2019 challenge-winning system (\fbkb) \cite{gretter2019}. It is seen that the proposed networks (albeit competitive w.r.t.\ the other participants in the Challenge \cite{Baur2019}) did not improve the state-of-the-art. Cautions are required in assessing the present relative comparison, insofar that the \fbkb\ is a multiple-classifier system,
exploiting several different feature sets as well as a number of different NN architectures at once. The best performing stand-alone DNN turns out to be the LSTM-W2V-M, shown in boldface in Table~\ref{tab:proposed_approaches_results}. It is observed that both the accuracy and the $D_{full}$ of the LSTM-W2V-M are close to the corresponding metrics yielded by the  \fbkb. All in all, the use of alternate textual feature turns out to be viable and capable of performances that are in line with the baseline, although the latter could not be improved. The latter consideration suggests that exploiting the reference grammar and the probabilistic LMs remains rewarding in facing the task. Nonetheless, all the proposed stand-alone approaches improved significantly in terms of $D_{full}$ over the best system presented in \cite{sokhatskyi2019} (that ranked 3rd in the 2019 challenge), in spite of the latter exploiting also an underlying LM (beside the word embeddings). Furthermore, three of the proposed DNNs (LSTM-W2V-M, FFN-BOW-WC, LSTM-BERT) improved the $D_{full}$ over the best system presented in \cite{russell2019} (that ranked 2nd in the 2019 challenge), in spite of the latter exploiting the reference grammar besides Word2Vec. Noticeably, LSTM-W2V-M yielded a $8.79\%$ relative $D_{full}$ improvement over \cite{russell2019}. 

\begin{table} [h]
\begin{center}
\begin{tabular}{c c c c c}
\hline
{Model} & {$D_{full}$} & {Accuracy (\%)} & {F1}\\
\hline
LSTM-W2V & 5.65 & 86.1 & 0.88\\
LSTM-W2V-L & 4.92 & 86.1 & 0.88\\
\textbf{LSTM-W2V-M} & \textbf{6.19} & \textbf{87.3} & \textbf{0.88}\\
FFN-BOW-WC & 5.97 & 86.0 & 0.88\\
FFN-BOW-TFIDF & 5.59 & 85.3 & 0.88\\
LSTM-BERT & 6.04 & 85.8 & 0.90\\
\hline
\fbkb & 6.34 & 87.5 & 0.92\\
\hline
2nd-best \cite{russell2019} & 5.61 & n/a & 0.91\\
3rd-best \cite{sokhatskyi2019} & 5.43 & n/a & 0.91\\
\hline
\end{tabular}
\caption {Stand-alone DNNs: results on {\em TestSet}.}
\label{tab:proposed_approaches_results}
\end{center}
\vspace{-1 cm}
\end{table}

The subsequent experimental round revolved around the multiple DNNs systems, investigating whether different features/models could effectively combine and complement each other. In the following we will write PJ-1\fbkb-Dfull to represent the pseudo-joint (PJ) combination between the best network in the \fbkb\  multiple-classifier system and the best DNN (among those evaluated in Table~\ref{tab:proposed_approaches_results}) in terms of $D_{full}$, and PJ3-3\fbkb-F1 to denote the pseudo-joint combination of the 3-best \fbkb\ networks and the 3-best proposed DNNs in terms of F-measure. Model selection was carried out over a number of combinations of subsets of the 1-best/3-best/6-best Challenge-winning network(s) reviewed in Section \ref{subsec:FBKsys}, and the 1st/2nd/3rd best network(s) proposed in Sections \ref{subsec:LSTM_W2V} and \ref{subsec:MLP_BOW}. All available evaluation metrics were considered in the model selection process. In short, it turned out that the best results yielded by the pseudo-joint  combination technique relied on the 1-best DNN proposed in Sections \ref{subsec:LSTM_W2V} and \ref{subsec:MLP_BOW} and the 1-best  \fbkb\  network according to the $D_{full}$ (PJ-1\fbkb-Dfull), as well as on the 3-best models proposed in Sections \ref{subsec:LSTM_W2V} and \ref{subsec:MLP_BOW} and the 3-best  \fbkb\  networks according to the F-measure (PJ3-3\fbkb-F1). As for the mixtures of experts, it was observed that the top-notch performances were achieved by the 3-best  \fbkb\  networks according to the F-measure, combined with the 1-best (MIX-3\fbkb-F1) or the 2-best (MIX2-3\fbkb-F) DNNs presented in Sections \ref{subsec:LSTM_W2V} and \ref{subsec:MLP_BOW}, respectively. The gating DNN selected for the MIX-3\fbkb-F1 model was a 7-layers architecture with 4, 175, 75, 90, 50, 75, and 4 neurons per layer. The gating DNN selected for the MIX2-3\fbkb-F1 model had 7 layers, as well, having 5, 512, 126, 256, 126, 126, and 4 neurons per layer.

\begin{table}[h]
\begin{center}
\begin{tabular}{c c c c c}
\hline
{Model} & {$D_{full}$} & {Accuracy (\%)} & {F1}\\ 
\hline
PJ-1\fbkb-Dfull & 6.82 & 87.9 & 0.89 \\
PJ3-3\fbkb-F1 & 7.00 & 87.1 & 0.89\\
MIX-3\fbkb-F1 & 7.56 & 89.1 & 0.92\\
\textbf{MIX2-3\fbkb-F1} & \textbf{8.08} & \textbf{88.7} & \textbf{0.92}\\
\hline
\fbkb & 6.34 & 87.5 & 0.92\\
\hline
\end{tabular}
\caption {Mixtures of DNNs: results on {\em TestSet}.}
\label{tab:combination_results}
\end{center}
\vspace{-1 cm}
\end{table}

Results are reported in  \autoref{tab:combination_results}. It is seen that the mixtures of experts yielded the highest $D_{full}$, Accuracy, and F-Measure on the data of the 2019 challenge to date. In particular, the MIX2-3\fbkb-F1 resulted in a $27.44\%$ relative $D_{full}$ increase over \fbkb, and in a $9.60\%$ relative error rate reduction w.r.t. \fbkb. The combination based on the pseudo-joint approach proved effective, as well. In fact, the PJ3-3\fbkb-F1 yielded a significant improvement over \fbkb\ in terms of $D_{full}$, while PJ-1\fbkb-Dfull improved the baseline in terms of Accuracy. The outcome of the experiments pinpoints the fact that the different features used and the corresponding stand-alone DNNs do actually model diverse information on the linguistic phenomena at hand, fruitfully complementing the grammar-specific knowledge exploited by the \fbkb\ models.

\section{Conclusions}
\label{sec:conclusion}
%Relying on the  FBK speech recognition system and on the FBK DNN-based system for the third Spoken CALL Shared Task challenge, 

%The paper investigated several text representations that are suitable to DNNs and LSTM networks which, in turn, were trained on such representations to accomplish the \df{2019 spoken CALL task challenge}.

In their stand-alone versions, the DNNs proved competitive w.r.t.\ the state-of-the-art, improving over previous attempts to use NNs on word embeddings for the 2019 challenge. Both the techniques proposed for combining multiple DNNs achieved significant improvements over the winner of the 2019 challenge, yielding the highest values of the metrics for the task to date. Results confirm that different text representations and different DNNs may actually capture diverse facets of the linguistic phenomena at hand, complementing each other effectively. In future we aim to address more complex tasks in speech scoring (e.g., \cite{icassp2019,lrec2020}), and to extend the multiple classifiers so as to consider also DNNs experts trained on acoustic features.

\bibliographystyle{IEEEtran}

\bibliography{refs}

\end{document}